\documentclass{article} %
\usepackage{collas2024_conference,times}
\usepackage{easyReview}

\usepackage{dsbda-style}

\usepackage{booktabs}
\usepackage{adjustbox}
\usepackage[normalem]{ulem}
\usepackage{placeins}
\usepackage{wrapfig}
\usepackage{floatrow}
\usepackage{caption}
\usepackage{subcaption}

\setreviewsoff

\newcommand{\norm}[1]{ \left\| #1 \right\| }

\usepackage{amsmath,amsfonts,bm}

\def\eqref#1{equation~\ref{#1}}

\def\1{\bm{1}}

\DeclareMathAlphabet{\mathsfit}{\encodingdefault}{\sfdefault}{m}{sl}
\SetMathAlphabet{\mathsfit}{bold}{\encodingdefault}{\sfdefault}{bx}{n}

\DeclareMathOperator*{\argmax}{arg\,max}

\usepackage{hyperref}
\hypersetup{
    colorlinks=true,
    linkcolor=red,
    filecolor=magenta,
    urlcolor=blue,
    citecolor=purple,
    pdftitle={Overleaf Example},
    pdfpagemode=FullScreen,
    }

\title{POWN: Prototypical Open-World Node Classification}

\author{Marcel Hoffmann   \\
University of Ulm\\
Germany \\
\texttt{marcel.hoffmann@uni-ulm.de} \\
\And %
Lukas Galke \\
MPI for Psycholinguistics \\
Nijmegen, Netherlands\\
\texttt{lukas.galke@mpi.nl} \\
\And %
Ansgar Scherp \\
University of Ulm \\
Germany \\
\texttt{ansgar.scherp@uni-ulm.de}
}

\collasfinalcopy %

\begin{document}

\maketitle

\begin{abstract}
We consider the problem of \textit{true} open-world semi-supervised node classification, in which nodes in a graph either belong to known or new classes, with the latter not present during training. 
Existing methods detect and reject new classes but fail to distinguish between different new classes. 
We adapt existing methods and show they do not solve the problem sufficiently. 
We introduce a novel end-to-end approach for classification into known classes and new classes based on class prototypes, which we call Prototypical Open-World Learning for Node Classification (POWN). 
Our method combines graph semi-supervised learning, self-supervised learning, and pseudo-labeling to learn prototype representations of new classes in a zero-shot way. 
In contrast to existing solutions from the vision domain, POWN does not require data augmentation techniques for node classification. 
Experiments on benchmark datasets demonstrate the effectiveness of POWN, where it outperforms baselines by up to $20\%$ accuracy on the small and up to $30\%$ on the large datasets. 
Source code is available at \url{https://github.com/Bobowner/POWN}.
\end{abstract}

\section{Introduction}

\label{sec:introduction}
Node classification is the task of assigning labels to nodes of a graph based on information such as the node's features and the structure of the neighborhood.
A typical application is assigning topics to papers in a citation graph.
A common assumption in supervised node classification is that all classes in the test set were also part of the training set~\citep{GCN, GAT, Graph-MLP}, known as closed-world assumption.
Considering a real-world node classification setting, this assumption does not necessarily hold true.
For instance, in a co-purchase graph in e-commerce, vendors add products from new categories at any time.
In citation graphs, new research topics may arise that do not fit into the predefined categories, or in social networks, new communities may form discussing new topics.
In these scenarios, the unlabeled graph can be collected nearly for free, but acquiring labels is expensive as it involves human annotators.

In an open-world setting, methods need to deal with the appearance of new classes. 
This entails problems of out-of-distribution detection and generalization.
There are two kinds of strategies to handle the new classes in this setting: 
The model can either reject and deny to classify new classes~\citep{Open_WGL, OWGL_IJCNN, Galke_LGL}, \ie become a \emph{robust open-world} model, or the model integrates the new classes and classifies new instances in a zero-shot way~\citep{OpenCon, ORCA}, \ie become a \emph{true open-world} model.
Following the latter, we formulate the task of \emph{true open-world semi-supervised node classification}. 
Given a graph $G=(V,E)$, with nodes $V$ and edges $E$, a set of labeled nodes $V_l \subset V$ with known classes, and a set of unlabeled nodes $V_u \subset V$, such that $V = V_l \cup V_u$ and $V_l \cap V_u = \emptyset$.
The task is to classify the unlabeled nodes $V_u$ into all classes, including new ones, as $V_u$ may contain nodes of known and new classes.
It requires the model to be a strong classifier on the known classes, while being able to discover and classify new classes, too.

A similar problem has been studied by \citet{OpenCon} and \citet{ORCA} on images in computer vision.
However, these methods heavily rely on image augmentation \remove{strategies} to produce positive and negative samples for self-supervised contrastive learning.
\add{This renders them not directly applicable to node classification due to the non-\iid\xspace nature of each sample, \ie modifying edges of a node to change its class influences also the neighboring nodes.}

Existing works on open-world node classification~\citep{Open_WGL, OWGL_IJCNN, Galke_LGL} are limited as they only detect the presence of new classes and reject them.
Thus, they cannot distinguish between multiple new classes, \ie do not learn to classify them.

To address the true open-world semi-supervised node classification problem, we propose a novel end-to-end prototypical open-world learning method for node classification (POWN).
It is not only able to distinguish between known and new classes but can also classify nodes in the new classes in a zero-shot way without prior knowledge of the classes, \ie semantic class descriptions such as a class name~\citep{Gzeroshot1}.
The proposed POWN extends the work by~\citet{OpenCon}, a method designed for vision, by supporting the graph structure and avoiding the strong dependence on data augmentation.
POWN combines graph semi-supervised learning with self-supervised learning and pseudo-labels from label propagation~\citep{labelProp} to learn prototype representations for known and new classes.
It models each class by a prototype in the embedding space and assigns labels to nodes based on the distance to the closest prototype.
The node and prototype representations are learned by three loss functions: a supervised loss to learn representations of the labeled nodes, an unsupervised loss based on Deep Graph Infomax (DGI)~\citep{DGI} to learn representations of the unlabeled nodes, and a loss based on pseudo-labels to assign the unlabeled nodes to a prototype.
The pseudo-labels are assigned by label propagation where the edges are weighted by the distance to the closest prototype in the embedding space. 
This combination of label propagation on the edges and the distance of the embedding to the prototype bridges the gap between the graph topology and the embedding space for the pseudo-label assignment.

Our experiments confirm that POWN effectively tackles the true open-world node classification setting. 
It is able to maintain the accuracy on the labeled classes, while successfully outperforming the baselines by up to $20\%$ in accuracy on all classes, \ie known and new classes together. 
In summary, our contributions are 

\begin{itemize}

    \item We formalize the task of true open-world semi-supervised node classification. 
    For this task, the model has to be a strong classifier on the known classes, while being able to distinguish known and new classes, and learn meaningful representations without access to any labels.
    \item We propose the first true open-world semi-supervised node classification method for graphs called POWN.
    The method is trained end-to-end.  
    \item Experiments on six benchmark datasets show the performance of our method. We outperform all baselines based on GCN~\citep{GCN}, DGI~\citep{DGI}, spectral clustering~\citep{spectral_clust}, and OpenWGL~\citep{Open_WGL}, especially on large graphs.
    \item Ablation studies provide detailed insights into the embeddings and show the robustness to hyperparameter selection of the new method.
\end{itemize}

\section{Related Work}
\label{sec:relatedwork}

We introduce graph neural networks and self-supervised graph learning, which focuses on learning good representations solely based on the node features and edges.
We discuss open-world learning and the limitations of current literature. 
Finally, we describe graph few-shot learning, which uses some labeled data points for learning new classes. %

\paragraph{Graph Neural Networks and Graph Self-Supervised Learning}

Graph neural networks (GNNs) aggregate node representations by message-passing over the edges of the graph.
Among the most prominent GNNs are GCN~\citep{GCN}, GAT~\citep{GAT}, and GraphSAGE~\citep{GraphSage}.
GNNs became the standard model in (semi-)supervised node classification, where only a few labels are provided as training instances for each class.

In contrast to (semi-)supervised node classification, the goal of graph self-supervised learning is to acquire meaningful representations of the nodes without relying on any predefined labels~\citep{Graph_SSL_survey}.
During training, graph self-supervised learning models predict supervision signals computed only from the graph itself.
Depending on how these supervision signals are generated, these graph self-supervised learning methods can be categorized into generation-based, auxiliary property-based, contrast-based, and hybrid methods~\citep{Graph_SSL_survey}.
Generation-based methods train the model to reconstruct some part of the graph, \eg node features~\citep{Graph_gen_feat_andclust} or graph structure~\citep{Graph_gen_struct}.
Auxiliary-based methods define selected properties of the graph and train a model to predict them, \eg node degree or cluster indices~\citep{Graph_gen_feat_andclust}.
Contrast-based methods maximize the mutual information between two corrupted versions of the same node~\citep{DGI, GraphInfoClust, HTC_graph_contrast}.
A popular representative is Deep Graph Infomax (DGI)~\citep{DGI}.
It trains a model to predict whether a node or its corrupted version belongs to an aggregated vector representation of the graph, which is computed by averaging the node embeddings.
Expanding on DGI, \citet{GraphInfoClust} introduced an additional cluster-level similarity to the global aggregated vector of DGI and optimized both.
\citet{HTC_graph_contrast} used multiple autoregressively generated subgraphs as positive samples for the contrastive loss in graph classification.
Hybrid methods combine multiple objectives.
Two representatives are GPT-GNN~\citep{GPT-GNN}, which is simultaneously trained for feature-based and structure-based graph generation, and GMI~\citep{GMI}, which maximizes the mutual information among node neighbors while minimizing the generation error on edge reconstruction.

\paragraph{Open-World Learning}

The (semi-)\-supervised learning setting is based on the closed-world assumption, \ie each class observed during test time has been present during training~\citep{LML_book, GCN, Open_WGL}.
Settings that drop the closed-world assumption are denoted as \emph{open-world} settings.
In the open-world setting, there are two strategies to handle the new classes in the test data.
The model either rejects and refuses to classify new classes to become a \emph{robust open-world} model, or the model integrates new classes and classifies new instances in a zero-shot way to become a \emph{true open-world} model.
So far, research has focused mainly on the %
robust open-world setting~\citep{open_set_collas, General_OWML_Survey, Open_WGL, OWGL_IJCNN,Galke_LGL_IJCNN}.

There are two exceptions coming from the computer vision domain.
One is ORCA~\citep{ORCA}, which exploits the different learning speeds of known versus new classes by an adaptive margin mechanism and introduces a pseudo-label mechanism based on pairwise similarity, \ie assigns the same pseudo-label to close samples.
The other is OpenCon~\citep{OpenCon}, which learns representations based on three contrastive loss functions.
The supervised loss $\mathcal{L}_S$ is computed on the labeled dataset, pushing labeled samples close together in the embedding space.
The unsupervised loss $\mathcal{L}_U$ is computed on the unlabeled dataset, pushing similar examples together.
The third one is a supervised loss $\mathcal{L}_P$ based on the pseudo-labels determined by the closest class prototype.
The total loss is computed by a weighted sum of the three loss functions.
Both methods require data augmentation to generate positive and negative samples in contrastive learning.
Data augmentations as they are done for images, \eg rotation, translation, and clipping~\citep{SimCLR}, cannot be applied to graphs.
Although there are data augmentation strategies for graphs~\citep{DropEdge}, they are not suitable for creating explicit positive and negative samples for node classification.

In contrast to images, the work on graphs is limited to the robust open-world setting~\citep{Galke_LGL, OWGL_IJCNN, Open_WGL}.
OpenWGL~\citep{Open_WGL} trained a variational graph auto-encoder to learn embeddings, optimized to increase the uncertainty of new classes for the model and reject to classify nodes for which the prediction is highly uncertain.
\citet{Galke_LGL} applied Deep Open Classification~\citep{DOC} from the text domain to graphs by weighting the loss function to account for the class imbalance in citation graphs.
\citet{OWGL_IJCNN} proposed a meta-method for aggregating confidence scores combined with a weakly-supervised threshold detection method to reject new classes.
\citet{OWAGL} focused on active learning and thus require a human oracle to label a set of nodes under a given budget.
This way, they not only rejected but also learned new classes, similar to our work.
However, \citet{OWAGL} shift the focus towards determining a set of good nodes given to the annotator and relaxes the open-world scenario, where in general no labels are provided for new classes.
Overall, existing open-world learning approaches focus on identifying and excluding new classes, rather than classifying objects into them, or they rely on human annotators to handle new classes.

\paragraph{Graph Few-Shot Learning}

In graph few-shot learning, the model has to learn classes based on a few labeled nodes per class~\citep{GPN, Meta-GNN}, \eg three or five.
Most graph few-shot learning models use meta-learning where the objective is to learn a few-shot learner from a large number of labeled classes.
A prominent representative is Graph Prototypical Networks~\citep{GPN}. 
It learns to derive prototypes from a few labeled samples based on a node encoder and a node valuator, which assigns a weight to each node.
The prototype is the weighted mean of the samples from its class.
Graph Prototypical Networks has been extended to Geometer~\citep{Geometer} for the class incremental graph few-shot learning setting by adding geometry-motivated loss functions to the prototypes, \ie supporting a uniform distribution of the prototypes and maximizing the distance between prototypes.
Other methods, \eg Meta-GNN~\citep{Meta-GNN}, avoid using prototypes by applying meta-learning directly to the classifier.
Graph few-shot learning differs from our setting since we assume to encounter new classes without any labeled nodes in the training set.
In few-shot learning, a few labeled nodes are available, while in zero-shot learning the models have no access to training data of specific classes.
Thus, in zero-shot learning, prior knowledge is often required such as a semantic description of the class~\citep{Gzeroshot1}, \eg the class name ``biology''.
In our approach, we do not have such a requirement.

\section{Problem Formalization}
\label{sec:problemstatement}
Based on the discussions in the related work, we formulate the true open-world node classification task.
Given a graph $G=(V,E)$ and a set of classes $Y$, where $V$ is the set of nodes and $E$ is the set of edges.
The nodes are separated into a set of labeled nodes $V_l \subseteq V$  and a set of unlabeled nodes $V_u \subseteq V$.
The nodes in $V_l$ are from known classes $Y_k \subseteq Y$, while the nodes in $V_u$ need to be classified into known classes $Y_k$ as well as new classes $Y_n$.
For the latter, no instance has been observed in the training data $V_l$, \ie $Y_n \not\subset Y_k$.
The task is to determine a model $f$ that maintains performance on the known classes $Y_k$, while also correctly classifying the new classes $Y_n$ in a zero-shot manner.
To excel in this task, a model needs to unify three characteristics:
It needs to be a strong classifier of the known classes while distinguishing known and new classes, and is learning meaningful representations without access to any labels. 

\paragraph{Assumptions}
We assume to know the number of existing classes.
The number of potential new classes is either known from the domain or can be estimated by existing algorithms~\citep{n_class_estimate}.
Experiments with an estimated number of classes are provided in Appendix~\ref{app:number_of_class_estimate}.
We assume a transductive setting where the full graph structure and node features are available for training.
Such a transductive setting is relevant for many real-world scenarios, where the graph can often be collected easily but human annotations are expensive. 
Furthermore, we assume homophilic graphs, \ie connected nodes likely belong to the same class.
If a domain is homophilic, this property holds for known and new classes.
This is the foundation for the label propagation of pseudo-labels.
Many real-world graphs are homophilic, \eg citation graphs, social networks, or co-purchase graphs~\citep{GCN}.

\section{Prototypical Open-World Node Classification (POWN)}
\label{sec:methods}

\begin{figure}
    \centering
    \includegraphics[width=0.8\textwidth]{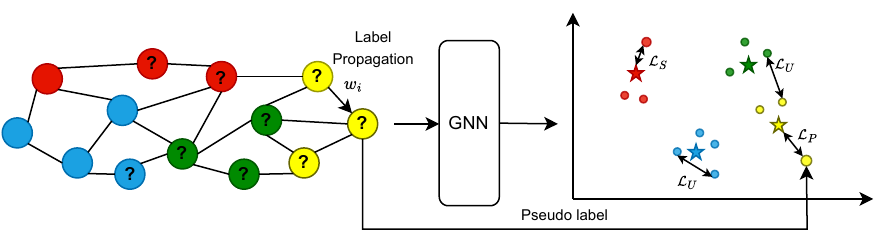}
    \caption{Overview of the losses in POWN. The input graph has four classes, where two are known (red and blue) and two are new (orange and green). The nodes without a question mark are nodes in $V_l$ and the nodes with a question mark are from $V_u$. Stars represent the prototypes in embedding space, and $w_i$ is the label propagation weight.}
    \label{fig:pipeline}
\end{figure}

Our proposed method, POWN, combines graph semi-supervised with self-supervised learning and pseudo-labels from label propagation to learn prototype representations for known and new classes.
An illustration of POWN is presented in Figure~\ref{fig:pipeline}.
Similar to the approach of \citet{OpenCon}, we use three different loss functions for different parts of the data.
The supervised loss $\mathcal{L}_{S}$ is computed on the labeled part of the data $V_l$, the unsupervised loss $\mathcal{L}_{U}$ is computed on the unlabeled part of the data $V_u$, and a pseudo-label loss $\mathcal{L}_{P}$ is computed on a subset of the unlabeled data $V_n$, where a new class can be determined with high confidence.
\add{The supervised and pseudo-label losses could be combined into one loss, but having them separate allows for applying different weights during training depending on the dataset.}

\subsection{Prototypes and Soft Prototype Membership}
Consider a set of prototypes $P$.
Each prototype $p_y \in P$ represents a (potential) class. 
We compute the embedding of each node $v$ by $z=f(x)$, where $f$ is some encoder, \eg a graph neural network.
We treat each prototype as an additional vector of learnable parameters of the neural network. 
The probability that a node $v$ belongs to prototype $p_i$, \ie class $y_i$, is computed by:
\begin{equation}
    \mathbf{p}_i(p_i | v) = \frac{\exp(-d(z,p_i)/\tau)}{\sum\limits_{ p_l \in P } \exp(-d(z,p_l)/\tau)},
    \label{eq:proba}
\end{equation}
where 
$d$ is some distance metric, 
$P$ the set of prototypes, and $\tau$ a temperature parameter.

\subsection{Prototype-based Representation Learning}
We compute different loss functions on different parts of the data.
The general aim is to minimize the distance between a prototype vector and the embeddings of the nodes belonging to the prototype's class by using the negative log-likelihood loss function:

\begin{equation}
    \mathcal{L}_\mathrm{nll}(V', P') = \frac{1}{|P'||V'|} \sum \limits_{p_i \in P'}\sum \limits_{v \in V'_i } -\log(\mathbf{p}_i(p_i | v)),
 \,\,\mathrm{where}\,
    V'_i = \{ v \in V' \mid \operatorname{label}(v) = i \ \}
\label{eq:prot_loss}
\end{equation}
$\mathbf{p}_i(p_i | v_j)$ is the probability computed in Equation~\ref{eq:proba}, with $P' \subseteq P$ a subset of the prototypes and $V' \subseteq V$ a subset of the nodes. 
The set $V'_i$ is a subset of $V'$ consisting of the vertices with the (pseudo-)label of class $i$, \ie $\operatorname{label}(v) = i$ means that the node $v$ belongs to class $i$.
Both, $V'$ and $P'$ are parameters that are defined later by the respective loss function.

To learn compact representations for the labeled data, we use the labels of the nodes in $V_l$ to minimize the distance between each node in $V_l$ and its corresponding prototype.
This is achieved by optimizing the supervised loss $\mathcal{L}_S = \mathcal{L}_\mathrm{nll}(V_l, P_k)$, where $P_k$ is the set of prototypes corresponding to a known class.   

Models for the image domain often combine data augmentation with contrastive learning to obtain good representations of unlabeled data~\citep{SimCLR, PCL}.
Creating positive samples by data augmentation for node classification is difficult~\citep{Graph_SSL_survey} because of the non-\iid\xspace property of the samples.
Therefore, we use the Deep Graph Infomax (DGI) loss, which only requires creating negative samples. 
For applying DGI, two additional functions are needed, a corruption function that modifies node features and a summary function to compute an embedding of the whole graph by aggregating all node embeddings.
For both, we follow the results of DGI and shuffle node features as a corruption function and take the mean of the node embeddings as a summary function.
Given the corruption and summary functions, the DGI loss for the unlabeled data $V_u$ results in:
\begin{equation}
    \mathcal{L}_U = \frac{1}{|V_u|} \sum \limits_{v_i \in V_u} -\left(\log\left(D(z_i,s)\right) + \log\left(1 - D(\tilde{z}_i, s)\right)\right),
\end{equation}
where $s$ is the summary vector of the graph, $\tilde{z}_i$ is the corrupted version of $z_i$, and $D$ is a linear binary classifier that is trained to distinguish between samples that belong to the summary $s$ and samples that do not. 

Subsequently, we split the unlabeled nodes $V_u$ in a set $V_n \subset V_u$ and $\overline{V_n} = V_u \setminus V_n$ based on their distance to known-class prototypes.
The subset $V_n$ contains all unlabeled nodes which likely belong to new classes and $\overline{V_n}$ contains all unlabeled nodes which likely belong to known class.
We define the subset $V_n$ of unlabeled nodes $v_i \in V_u$ by 
\begin{equation}
    V_{n} = \{v_i \in V_u | \max \limits_{y \in Y_k}\{ \langle\,p_y, z_i\rangle) \} < \gamma \},
\end{equation} 
where $v_i$ belongs with high confidence to a new class, \ie
its embedding $z_i$ is highly different from a prototype of one of the known classes.
The threshold $\gamma$ is determined on the labeled data $V_l$ such that $q$ percent, \eg $q=90\%$, of the labeled data is above the threshold.
This procedure implements a simplified OOD detection based on the method of \citet{EntropicOOD} to obtain high likelihoods that nodes in $V_n$ belong to a new class.
For each node $v_i \in V_n$, we determine the closest prototype based on cosine distance and assign the pseudo-label $\hat{y}$ of the prototype to the node, \ie $\hat{y} = \argmax_{y \in Y_n}(\langle\,p_y, z_i\rangle)$.
We calculate the loss with respect to the pseudo labels as a variant of Equation~\ref{eq:prot_loss} by $\mathcal{L}_P = \mathcal{L}_{nll}(V_n, P_p)$, where $P_p$ is the set of prototypes to which at least one node \add{of $V_n$} has been assigned to by the pseudo-label method.
We assume that there are new classes in $V_n$, use $V_n$ to assign pseudo labels and calculate the loss $\mathcal{L}_P$, while nodes in $V_n^c$ are only affected by the unsupervised loss.

\subsection{Assignment and Propagation of Pseudo Labels}
To update the prototypes for new classes, we assign pseudo-labels to the unlabeled nodes.
Our pseudo-label method is based on the homophily assumption of the label-propagation algorithm~\citep{labelProp}.
To obtain the final labels, we perform three steps:
First, we assign to the nodes in $V_n$ the pseudo label of their closest prototype as described in the previous section.
The pseudo-labeled nodes serve as seeds for the second step, our weighted label-propagation algorithm, where each edge $(v_i, v_j)$ is weighted by the distance of \replace{$v_i$}{$v_i \in V$} to its closest prototype.
This inhibits the label flow from uncertain nodes.
Thus, the weighting combines the graph topology with the embedding space.
Given two nodes $v_i, v_j$ and the edge $(v_i, v_j)$, %
 the weight is  
\begin{equation}
    w_{j} = 1  / (\norm{z_j - p_c}) \,,
\label{eq:weights}
\end{equation}
where $z_j$ is the embedding of $v_j$ and $p_c$ is the prototype that is closest to $v_i$ in the embedding space.
These weights enhance the propagation of more confident labels (close to a prototype) compared to less confident ones.
After $l$ hops of label propagation, we apply a softmax to the output and obtain a probability distribution for each node.

In the third step, we compute the Shannon entropy of the label distribution of each pseudo-labeled node and remove the $10\%$ with the highest entropy.
This ensures that only highly confident~\citep{EntropyScore} pseudo-labels are used for the loss update, and nodes that either have a too heterogeneous neighborhood for the label propagation to perform well or have not been affected by the label propagation (because $l$ was too low) are removed. 

At the beginning of the training, the edge weights are noisy and small since the mean distance of the embeddings to a prototype is high. 
This inhibits the effect of label propagation and avoids propagating wrong labels of randomly assigned prototypes.
As the training continues, the weights increase and facilitate the flow of the label propagation, leading to more confident labels and prototype assignments.
See Appendix~\ref{app:lp_weights} for an empirical analysis.

\subsection{Combined Loss Function}
The final loss function is the combination of the proposed losses plus a regularization, defined by
\begin{equation}
\label{eq:total_loss}
\mathcal{L} = \lambda \mathcal{L}_S + \mu \mathcal{L}_U + \nu \mathcal{L}_P + \kappa \mathcal{R} \,,
\end{equation}

where $\lambda$, $\mu$, and $\nu$ are the weights of the loss functions.
$\mathcal{R}$ is a regularization term consisting of the sum of the entropy regularization of \citet{ORCA} and the maximal negative pairwise distance between the prototypes of \citet{Geometer} with $\kappa$ as the weight of the regularization term. 
The entropy regularization is defined as the Kullback-Leibler divergence of the output class distribution and uniform distribution, preventing the output on the new classes from being too flat, \ie to collapse to one or only a few classes, and distributing the embeddings equally around the prototypes.
The pairwise prototype distance ensures that the prototypes are biased towards uniform distribution in the embedding space.
A formalization is given in Appendix~\ref{app:reg_term}.

\subsection{Complexity Analysis}
The computational complexity depends on the backbone model $f$.
Since we use a GCN, the complexity for $f$ is given by $\mathcal{O}(|E|h^{L-1}d)$~\citep{GCN}, where $|E|$ is the number of edges of the graph, $h$ the hidden dimension of the GCN, $L$ the number of layers, and $d$ the input feature dimension.
For the supervised loss, we have to compute the distances between each sample and each prototype, which can be done in $\mathcal{O}(|V||P|h)$.
For the unsupervised loss, we have to evaluate $f$ two times, resulting in the run-time complexity of GCN.
For the pseudo-labels, we do a constant number of iterations of label propagation, which is in $\mathcal{O}(|E||Y|)$, then we do the same computation as for the supervised loss.
We compute the pairwise distance of the prototypes for the regularization term, which scales with $\mathcal{O}(|Y|^2) = \mathcal{O}(|V|)$ since it holds that $|Y|^2 \ll |V|$ for most applications.
As these complexities are additive, the complexity of POWN stays linear w.\,r.\,t. the input dimensions $|V|$, $|E|$, and $d$.

\subsection{Summary}
We formulated the problem of true open-world semi-supervised node classification and proposed POWN, an open-world learning method to classify seen as well as unseen classes. 
We outlined how to combine information in the embedding space with graph topology to assign pseudo-labels for potentially new classes.
Next, we introduce the experimental apparatus to evaluate the performance of our method.

\section{Experimental Apparatus}
\label{sec:experimentalapparatus}

\extended{
We introduce the datasets and describe our experimental procedure, hyperparameter search, and measures.
}

\subsection{Datasets}
\label{sec:datasets}
To show that POWN can distinguish between the new classes, we ensure that the datasets have at least two new classes in the validation set and two new classes in the test set.
We use the citation graphs Cora, CiteSeer~\citep{CoraCiteSeer}, and OGB-arXiv~\citep{ogb-benchmarks} as well as the co-purchase graphs Amazon-Photo (Photo)~\citep{PitfallsGNNEval}, Amazon-Computers (Computers)~\citep{PitfallsGNNEval}, and the social network Reddit2~\citep{GraphSaint}.
Descriptive statistics of the datasets are shown in Table~\ref{tab:datasets}, \ie the number of nodes $|V|$, the number of edges $|E|$, the dimension of the features $d$, the number of classes $|Y|$, and the homophily $\mathcal{H}$.
The homophily $\mathcal{H}$, the tendency of connected nodes to share the same class, is measured by the class-insensitive homophily measure~\citep{class-insensitive-homophily}.

\begin{table}[!th]
    \centering
    \caption{Number of nodes $|V|$, edges $\lvert E \rvert$, features $d$, classes $\lvert Y \rvert$, and homophily measure $\mathcal{H}$ of the datasets.}
    \label{tab:datasets}
    \begin{tabular}{lrrrrr}
    \toprule
     & $\lvert V \rvert$ & $\lvert E \rvert$ & $d$ & $\lvert Y \rvert$ & $\mathcal{H}$\\
    \midrule
    Cora      & $2,708$ & $10,556$ & $1,433$ & $7$ & $0.766$\\
    CiteSeer   & $3,327$ & $9,104$ & $3,703$ & $6$ & $0.627$\\
    OGB-arXiv & $169,343$ & $1,166,243$ & $128$ & $40$ & $0.421$\\
    Photo & $7,650$ & $238,162$ & $745$ & $8$ & $0.772$\\
    Computers & $13,752$ & $491,722$ & $767$  & $10$ & $0.700$\\
    Reddit2 & $232,965$ & $23,213,838$  & $602$ & $41$  & $0.691$\\

    \bottomrule
    \end{tabular}
\end{table}

\subsection{Procedure}
\label{sec:procedure}

\paragraph{Node Split and Class Split}
As in the common node classification setting, we need to divide the nodes into train, validation, and test nodes, which are fixed across all experiments.
We use the Planetoid split~\citep{Planetoid} for Cora and CiteSeer and we employ the split of \citet{PitfallsGNNEval} for Photo and Computers.
For OGB-arXiv, we use the default split from \citet{ogb-benchmarks} and for Reddit2 the split of \citet{GraphSaint}. 
Details on the node split can be found in Appendix~\ref{app:node_split}.

Additionally, we need to split the classes into train, validation, and test classes to evaluate true open-world learning.
We split the classes into labeled and new classes, $Y = Y_l \overset{.}{\cup} Y_u$.
From the labeled classes, we split a third subset $Y_v \subset Y_l$ for validation.
The number of new classes is defined by the new class ratio $r$.
We partition the classes into three sets.
We split $r \%$ of the classes, which are used as new classes for testing and another $r\%$ of the classes, which are used for validation.
This results in the labeled node set $V_{l}$, which consists of $(1-2r)|Y|$ classes, for training, a test set $V_u$, which is equal to the test set of the dataset, containing all, \ie labeled and new, classes, and a validation set containing $r|Y|$ new classes and labeled classes.
Following the splits over nodes and classes, we obtain unlabeled nodes in the train set that stem from validation classes and test classes.
Test nodes come either from classes known during training or new classes.

\paragraph{Folds and Repeats}
Since the selection of classes may have a high impact on the overall performance, we use a cross-validation method to minimize potential threats to the validity of the results.
Given the ratio $r$ of new classes, we split the classes into $1/r$ folds.
We choose the new class ratio as $r=0.2$.
The experiment is repeated for each fold, where each fold once represents the new test classes and another fold represents the new validation classes.
Note that the set of training, validation, and test nodes is fixed all the time, only the labeled versus unlabeled classes are varied across the folds.

We repeat the whole procedure over all folds $5$ times for the smaller datasets, \ie Cora, CiteSeer, Photo, and Computers, resulting in $25$ overall runs.
We repeat the experiment two times for the larger graphs OGB-arXiv and Reddit2, resulting in $10$ overall runs each.
Details on the dataset folds can be found in Appendix~\ref{sec:ds_details}. 

\paragraph{Baselines}
We use POWN with a GCN~\citep{GCN} backbone for all experiments.
Since we are the first to explore true open-world semi-supervised node classification, there is no direct method we can compare to.
Therefore, we use and adapt existing methods to apply to our setting.
As a representative of common semi-supervised node classification methods, we select a GCN with an output layer equal to the number of labeled and new classes.
For unsupervised methods, we use DGI~\citep{DGI} embeddings, which we cluster by $k$-means and spectral clustering of the graph structure.
For both methods, we provide the number of classes as the number of clusters.
Furthermore, we extend the existing robust open-world learning method OpenWGL~\citep{Open_WGL} to the true open-world semi-supervised node classification setting.
It classifies the samples into the labeled classes and rejects all instances that do not belong to a labeled class.
We apply $k$-means clustering on the embeddings of all rejected instances to obtain a separation of new classes.
Details on the training procedure can be found in Appendix~\ref{appendix:hyperparams}.

\paragraph{Hyperparameter Optimization}
\label{sec:hyperparameteroptimization}
To ensure comparability, all compared methods use the same GCN backbone. 
We use the GCN hyperparameter values of \citet{HyperParams} for Cora and CiteSeer.
For Photo and Computer, we use the parameter values of \citet{PitfallsGNNEval}, and for OGB-arXiv, we use the hyperparameters of \citet{OWGL_IJCNN}.
On Reddit2, we use the hyperparameter values of \citet{GraphSaint}.
For the OpenWGL baseline, we have tuned the hidden size with possible values $\{16,32,64\}$ and a fixed learning rate of $0.001$. 
For the DGI baseline, we employ standard feature shuffle as a corruption function and standard mean aggregation as graph-level readout. 
For POWN, we tune the scalars $\lambda$, $\mu$, $\nu$ of the loss function, and $\kappa$ by Bayesian optimization. 
We select the hyperparameters with the highest average validation accuracy on all classes computed over all folds and repeats.
Further details on the procedure and final hyperparameter values can be found in Appendix~\ref{appendix:hyperparams}.

\subsection{Measures}
\label{sec:measures}

We use the evaluation measures of \citet{OpenCon,ORCA}. %
For the \textbf{known} classes, we compute the known-class accuracy as the ratio of correct predictions to all predictions.
For the \textbf{new} classes and \textbf{all} classes, the methods cannot determine which of the new classes belong to which label.
For this reason, we first solve an optimal assignment problem between the predicted and true labels using the Hungarian algorithm~\citep{HungarianAlg}.
Based on this assignment, we compute the \textbf{new} class accuracy and the \textbf{all} class accuracy, respectively.

\section{Results}
\label{sec:results}

\begin{table*}[!ht]
    \centering
    
    \caption{
    Mean accuracy with standard error across up to $5$ class folds and $10$ runs for each of the three cases: All classes (Top), Known classes (Center), New classes (Bottom). The best score per measure is marked in bold. Unsupervised methods DGI+$k$-means and Spectral clustering are not applicable for the supervised setting with known classes.}
    \begin{adjustbox}{width=\textwidth}

     \begin{tabular}{l|rrrr|rr}
    \toprule
     &   \multicolumn{4}{c|}{Small datasets} & \multicolumn{2}{c}{Large datasets} \\
     \midrule
     Method  & \multicolumn{1}{c}{Cora}  & \multicolumn{1}{c}{CiteSeer}  &  \multicolumn{1}{c}{Photo} &  \multicolumn{1}{c|}{Computers} &  \multicolumn{1}{c}{OGB-arXiv} &  \multicolumn{1}{c}{Reddit2}  \\
    \midrule
     \textbf{All classes w/ Hungarian Algorithm} &&&&&& \\
     GCN &  $54.48_{0.74}$  & $50.99_{0.92}$  & $52.21_{1.77}$  & $61.40_{2.75}$  & $37.98_{1.45}$  & $46.05_{2.19}$ \\

     DGI + $k$-means &   $38.85_{2.22}$  & $38.67_{2.57}$  & $25.87_{0.94}$  & $22.27_{0.80}$ & $28.16_{1.45}$ & $17.10_{0.96}$ \\
     Spectral clustering &  $ 29.61_{0.97}$ & $32.26_{0.67}$ & $27.11_{0.35}$ & $29.99_{0.54}$  & $24.24_{0.71}$ &  \multicolumn{1}{r}{OOM} \\
    OpenWGL + $k$-means &  $\mathbf{64.10_{2.25}}$  &  $50.38_{2.48}$  & $62.45_{0.94}$  & $51.18_{1.61}$  & $31.62_{1.72}$  & $15.93_{1.08}$ \\
    POWN (own) &  $61.28_{1.09}$  & $\mathbf{56.15_{1.56}}$  & $\mathbf{71.27_{1.44}}$  & $\mathbf{71.33_{1.84}}$   & $\mathbf{55.51_{3.26}}$ & $\mathbf{76.99_{2.19}}$ \\

    \midrule
    \textbf{Known classes (Fully Supervised) }&&&&&& \\
    GCN  & $\mathbf{95.19_{0.28}}$  &  $\mathbf{85.93_{1.43}}$  & $89.00_{1.67}$  & $\mathbf{91.99_{3.65}}$  &  $46.37_{4.04}$  &  $55.16_{2.34}$ \\

     OpenWGL + $k$-means  &   $61.74_{3.69}$  &  $38.18_{0.99}$  &  $60.06_{1.35}$ &  $36.93_{4.06}$  &  $27.04_{2.58}$  &  \: $7.48_{1.50}$ \\
      POWN (own)  & $87.95_{1.99}$  &  $75.45_{2.31}$ &  $\mathbf{91.05_{0.97}}$   &  $82.45_{2.91}$  &  $\mathbf{71.24_{2.60}}$ &  $\mathbf{92.52_{2.34}}$ \\
    \midrule
    
    \textbf{New classes w/ Hungarian Algorithm} &&&&&&\\
     GCN & $62.16_{1.82}$ &  $57.75_{1.60}$  & $65.24_{1.53}$ &  $\mathbf{69.37_{1.46}}$ & $42.82_{4.10}$ &  $ 55.01_{2.11}$\\
         DGI + $k$-means &   $41.97_{2.21}$ &  $41.17_{2.21}$ & $32.97_{3.81}$ & $40.40_{5.91}$ & $49.06_{4.10}$& $19.62_{1.42}$ \\
    Spectral clustering &   $39.22_{1.47}$ &  $37.70_{0.51}$ &  $36.68_{0.86}$ &  $56.51_{2.13}$ &  $37.81_{3.33}$ &  \multicolumn{1}{r}{OOM} \\

    OpenWGL + $k$-means &  $\mathbf{68.88_{3.22}}$  &  $\mathbf{58.88_{2.22}}$ &  $66.04_{2.48}$ &  $55.99_{4.36}$ &  $28.16_{1.29}$ &  $32.38_{1.36}$\\
     POWN (own) &   $63.29_{1.59}$ &  $57.29_{2.22}$ &  $\mathbf{70.76_{2.74}}$ & $65.61_{2.95}$ &  $\mathbf{56.11_{4.35}}$ & $\mathbf{64.42_{2.11}}$\\
     
    \bottomrule
    \end{tabular}
    \end{adjustbox}

    \label{tab:main_results}
\end{table*}

Table~\ref{tab:main_results} shows the results of the true open-world 
node classification setting.
We report all-class accuracy, known-class accuracy, and new-class accuracy along with their standard error over all class folds and runs.
We observe that POWN outperforms all baselines in terms of all-class accuracy on all datasets except Cora, where OpenWGL is the best-performing model.
In general, OpenWGL is the second-best model in terms of all-class accuracy.
In terms of known class accuracy, POWN loses up to $8\%$ on the small datasets compared to the supervised GCN, but improves the known-class accuracy on the large datasets.
We do not report known-class accuracy for the unsupervised methods since they cannot learn the mapping between the cluster ids and labels of the dataset, \ie their known class accuracy is simply random.
The best new-class accuracy depends on the dataset and is either achieved by GCN, OpenWGL, or POWN. 
However, POWN outperforms all methods on the large datasets by a wide margin of at least $18\%$ in the accuracy on all classes. 
Spectral clustering had an out-of-memory on  Reddit2 on a server with 2 TB of RAM.

\section{Discussion}
\label{sec:discussion}
\paragraph{True Open-World Semi-Supervised Node Classification}
Our results show that POWN effectively approaches the true open-world semi-supervised learning problem, \ie learns representation for nodes of known and new classes at the same time.
It improves the accuracy over all classes by up to $20\%$ on the small and up to $30\%$ on the large datasets over the best baseline.
We observe that the margin between POWN and the baselines increases on large datasets, which is the exact opposite trend of the performance of OpenWGL.

The positive effect of the dataset size on POWN's performance is attributed to two reasons. 
First, POWN's contrastive losses benefit from the larger amount of data. 
Second, in contrast to models relying mainly on labels, like GCN and OpenWGL, POWN learns from all parts of the data, even when there are no labels available.

POWN loses some accuracy on the labeled classes compared to the fully supervised GCN on the small datasets.
However, it is always better in terms of all-class accuracy since it also learns to separate known from new classes and uses unlabeled nodes more explicitly in its loss functions.
Nevertheless, GCN performs comparably well in capturing the structure of the new classes, especially on Computers. 
The homophily of the datasets and the message passing of GCN helps the model to group embeddings of the same class together, even without access to the labels.
Since our evaluation procedure takes the best mapping of the output labels to the classes using the Hungarian algorithm, GCN is a strong baseline in all measures.

The unsupervised baselines consistently have the lowest scores over all datasets and measures, showing that utilizing label information is crucial for learning meaningful representations.
In particular, the known-class accuracy becomes meaningless without supervised information.
Note that the all-class accuracy is not just the average of the known-class accuracy and the new-class accuracy. 
As defined in Section~\ref{sec:measures}, the \textbf{all-class} accuracy and \textbf{new-classes} accuracy are calculated based on the assignments obtained from the Hungarian algorithm to receive a mapping between the labels and clusters.
For this reason, the known-class accuracy can also be lower than the all-class accuracy where the Hungarian algorithm improves the accuracy by mapping the output labels to the true labels.

We have used homophilic datasets from different domains and of different sizes in our experiments.
We expect that our results do generalize to other homophilic datasets. 

\paragraph{Hyperparameter Sensitivity Analysis}
\begin{figure*}[!ht]
    \centering
    \includegraphics[width=0.9999\textwidth]{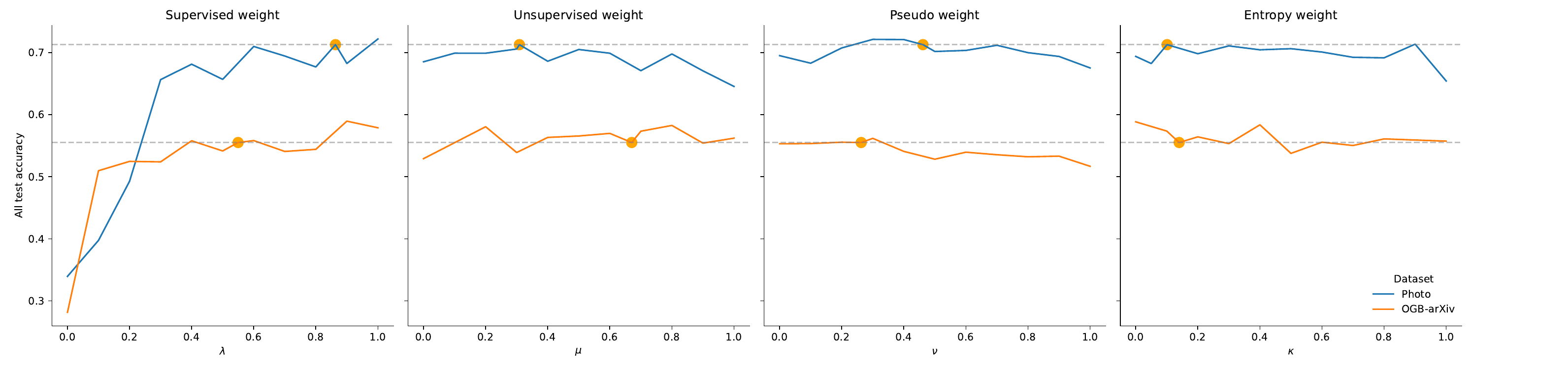}
    \caption{Test accuracy on all classes for variations of the hyperparameters $\lambda$, $\mu$, $\nu$, and $\kappa$ (figures from left to right). The orange dot marks the hyperparameters with the highest validation accuracy found by Bayesian search.}
    \label{fig:hp_sensitivity}
\end{figure*}

We conducted a hyperparameter sensitivity analysis to examine how POWN behaves for deviations from our hyperparameter values. 
We use the small dataset Photo and the large dataset OGB-arXiv.
We change POWN's loss weights $\lambda$, $\mu$, and $\nu$, and the entropy weight $\kappa$, while we keep all other values found by our hyperparameter optimization.
The results are presented in Figure~\ref{fig:hp_sensitivity}.
We observe that the model performance is quite robust for variations of the hyperparameters, except for the supervised loss weight $\lambda$ that has to be sufficiently high.
For OGB-arXiv, running Bayesian optimization for more iterations may have further improved the hyperparameters.
\add{Additional analysis of the temperature parameters can be found in Appendix~\ref{sec:temp_sens}.}

\paragraph{Ablation Study}

We run an ablation study to verify that all parts of the loss contribute to the performance.
We repeat the main experiment but remove one of the loss functions for each setting.
The results for Photo and OGB-arXiv are presented in Table~\ref{tab:ablation_study}, where each line leaves out one of the loss functions.

\begin{table}[!ht]
    \centering
    \normalsize	
    \begin{adjustbox}{width=0.6\textwidth}
    \begin{tabular}{c|ccc|ccc}
    \toprule
      &  \multicolumn{3}{c|}{Photo} & \multicolumn{3}{c}{ OGB-arXiv} \\
    \midrule
    & All & Known & New & All & Known & New \\
    \midrule
      w/o $\mathcal{L}_S$   & $33.95_{0.71}$ & \:\:---\:\:\: & \dashuline{$72.54_{0.40}$} & $28.16_{2.18}$ & \:\:---\:\:\: & $49.06_{5.65}$ \\
      
      w/o $\mathcal{L}_U$   &$68.51_{1.68}$ & \underline{$92.76_{0.66}$} & $70.18_{1.89}$ & $52.91_{2.49}$ & $69.09_{3.39}$ & $50.80_{2.73}$\\
      w/o $\mathcal{L}_P$   &$69.50_{1.99}$ & $90.43_{1.78}$ & $70.94_{2.90}$ & $55.31_{1.56}$ & \underline{$72.71_{2.01}$} & $55.46_{3.77}$\\
      w/o $\mathcal{R}$     &$69.38_{1.49}$ & $89.97_{0.93}$ & $65.62_{2.71}$&$54.84_{1.38}$ & $66.70_{1.89}$ & $53.27_{2.89}$ \\
      \midrule
      POWN  & $\mathbf{71.27_{1.44}}$ & $91.05_{0.97}$ & $70.76_{2.74}$ & $\mathbf{55.51_{3.26}}$ & $71.24_{2.60}$  & \dashuline{$56.11_{4.35}$}\\
    \bottomrule
    \end{tabular}
    \end{adjustbox}
    \caption{Ablation study: Accuracy on all, the known, and the new classes on the loss functions of POWN with the respective standard error.}
    \label{tab:ablation_study}
\end{table}

The combination of all loss functions gives the best result with respect to all-class accuracy. 
Leaving out a loss can improve the performance on some subset of the classes, \eg removing the supervised loss improves the performance on the new classes.
However, the performance on the known classes collapses, since there is no mapping of label indices to model output indices anymore.
\add{
Furthermore, the regularization term $\mathcal{R}$ contributes most to the new classes by preventing the embeddings to collapse around one or only a few prototypes.
Assuming the regularization term distributes the node representations among the prototypes, the unsupervised loss ensures that similar node representations are close to the same prototype.
The pseudo-label loss contributes to the overall performance by denoising the edges, \ie assigning high weights to homophilic and low weights to heterophilic edges.
It mainly improves the performance on all classes, where the edge denoising effect and the homophily of the graph lead to a better separation of known from new classes. 
It also improves the accuracy of the known classes, although it does not affect these prototypes.
This is due to the shared GNN encoder.}

\extended{
\paragraph{Variation of the Number of New Classes}
We examine the perfromance of POWN for variations of ratios of new classes compared to the baselines.
For this experiment, we gradually increase the ratio of new classes from $0.0\%$ to $100\%$ in $20\%$ steps and train and evaluate for each step.
The result is presented in Figure X.
}

\begin{figure}[!ht]
\floatbox[{\capbeside\thisfloatsetup{capbesideposition={right,top},capbesidewidth=0.5\textwidth}}]{figure}[\FBwidth]
{\caption{Left plot: $t$-SNE embeddings of GCN, right: POWN on the Photo dataset. Known classes are colored in gray, each new class in a different color.} \label{fig:tsne}}
{\includegraphics[width=0.5\textwidth]{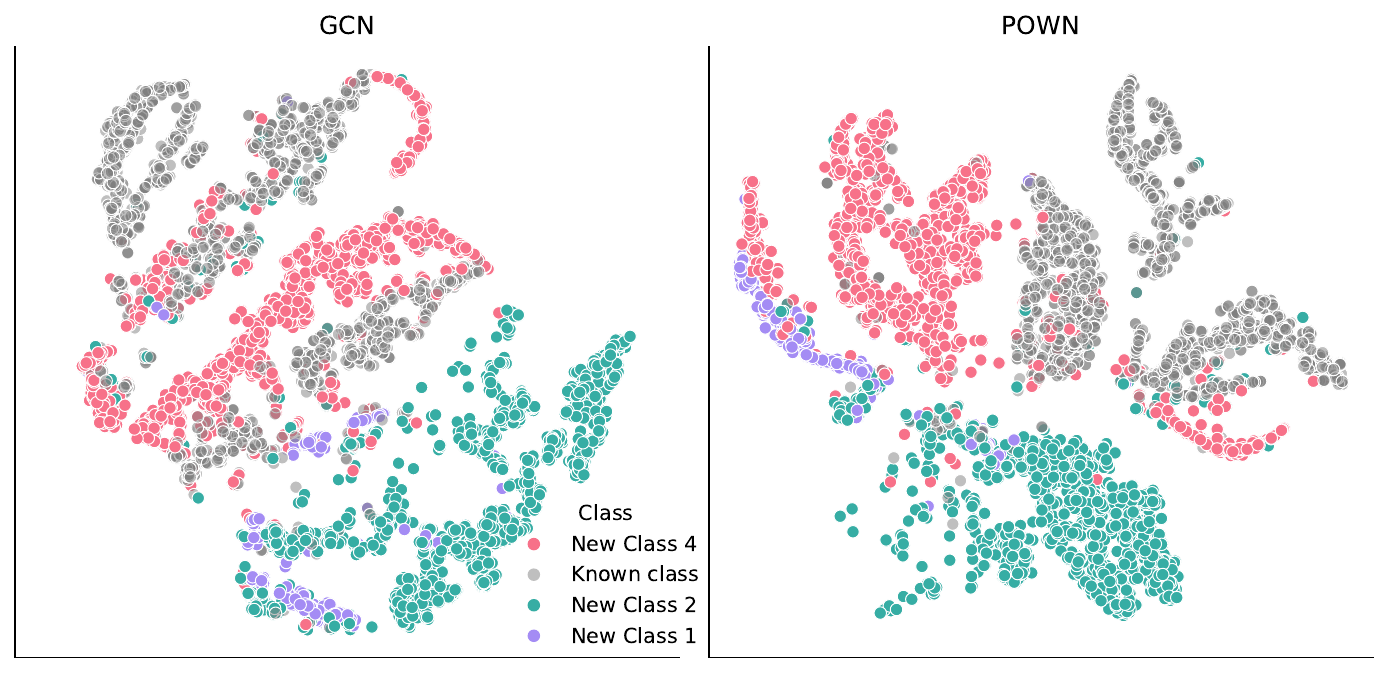}}
\end{figure}

\paragraph{t-SNE Embeddings}
For a qualitative assessment of the embeddings, we compare the $t$-SNE-reduced embeddings~\citep{t-SNE} of GCN and POWN on the unlabeled part $V_u$ of the Photo dataset.
The known classes are all shown in gray, while the new classes have different colors.
The embeddings are presented in Figure~\ref{fig:tsne}.
We see that POWN produces denser clusters on the known classes and new classes.

\paragraph{Estimating the \# of Classes}
For the main setting, we had assumed that
the number of classes $|Y|$ is known. 
To investigate the effect of this assumption, we ran the experiments with an unsupervised estimator for the number of classes \citet{n_class_estimate}, see details in Appendix~\ref{app:number_of_class_estimate}. 
The results show that the difference in performance on for all-classes is below $1\%$.

\section{Limitations}
\label{sec:threattovalidity}
Our method relies on the homophily assumption. %
Although many real-world graphs are homophilic, like the datasets we use in the experiments, the model will perform worse on heterophilic datasets.
POWN is inherently transductive due to the learning of prototypes from unlabeled data. However, for every application, the test data has to be available at some point. 
Thus, one can re-train POWN as soon as the data is available.
While POWN achieves remarkable performance on all classes combined, it falls behind the accuracy of a fully supervised GCN on the known classes on the small datasets.
The performance of the classification task highly depends on the specific class fold, since not all classes are equally difficult to distinguish from each other, leading to a high expected standard error w.r.t. the chosen classes for training, validation, and test. 
Therefore, we applied a cross-validation approach and averaged our results over three class folds on Cora and CiteSeer, four on Photo, and five on Computers, OGB-arXiv, and Reddit2. Additionally, we averaged the results over multiple repeats, \ie random seeds, per class fold.

\section{Conclusion}
\label{sec:conclusion}
\label{sec:futurework}
We address the problem of true open-world semi-supervised node classification.
We adapted existing methods from the literature to be suitable for the problem and show that they could not sufficiently solve it.
Therefore, we propose POWN, an end-to-end prototype-based open-world learning model that can effectively tackle the true open-world learning problem.
Our experiments show that POWN outperforms all baselines, especially on large graphs, and is robust to the hyperparameter selection.
Future work may investigate the effect of providing few, \eg one to five, examples for unlabeled classes to measure the difference to a labeled setting.
Furthermore, POWN can be extended to temporal graphs to see how the performance evolves if multiple iterations of POWN are applied.

\FloatBarrier

\bibliography{collas2024_conference}
\bibliographystyle{collas2024_conference}

\appendix
\label{appendix:supplementarymaterials}

\extended{
\subsection{Contrastive Loss Functions}

From literature various contrastive loss function to learn data and prototype representations are known~\citep{OpenCon, Geometer, Graph-MLP}. 
A major class is based on the following abstract formulation of the InfoNCE loss:

\begin{equation*}
    \mathcal{L}_{\phi}(\mathcal{P}, \mathcal{N}, \tau, x) = -\frac{1}{|\mathcal{P}(x)|} \sum_{z^+ \in \mathcal{P}(x)} \log \left(\frac{\exp{(z^T z^+ / \tau)}}{\sum_{z^- \in \mathcal{N}(x)} \exp{(z^T z^- / \tau)}}\right) 
\end{equation*}

where $x$ is a node, $z=f(x)$ the embedding of $x$ by a model $f$, $\mathcal{P}(x)$ the set of positive samples for $x$, $\mathcal{N}(x)$ the set of negative samples for $x$ and $\tau$ some smoothness parameter.
Note that the similarity $z^Tz^+$ is equivalent to cosine similarity if $z$, $z^+$ are normalized to length one.

Depending on the definition of $\mathcal{P}(x)$ and $\mathcal{N}(x)$ one gets different contrastive loss functions.
Choosing $\mathcal{P}(x)$ as the set of all samples with the same label leads to a supervised contrastive loss.
Choosing $\mathcal{P}(x)$ at similar samples, \eg augmented from the same sample, leads to a self-supervised contrastive loss. 
Especially for graphs you can receive the Graph-MLP~\citep{Graph-MLP} by choosing $\mathcal{P}$ as the neighbors of $x$.
The negative set $\mathcal{N}(x)$ is often chosen as all other samples.

\begin{equation*}
    \mathcal{L}_U =\sum_{x_i \in D_t} -\log \frac{\exp{(-\mathbbm{1}_{[i\neq j]} \gamma_{ij} d(f(x_i), f(x_j))/\tau)}}{\sum_{x_j \neq x_i }\exp{(-d(f(x_i), f(x_j))/\tau)}} 
\end{equation*}

\begin{equation}
     \gamma_{ij} \begin{cases}
= \alpha,& \text{node $j$ is the $r$-hop neighbor of node $i$} \\
 = 1-\alpha,& \text{node $j$ is not the $r$-hop neighbor of node $i$ }\\
\end{cases}
\label{eq_nc}
\end{equation}
with $\alpha > 0.5$ is a hyperparameter that depends on the homophily.
}

\newpage

\section{Regularization Term}
\label{app:reg_term}

The regularization term $\mathcal{R}$ consists of two parts based on \citet{ORCA} and \citet{Geometer}. 
It is defined by

\begin{equation}
    \mathcal{R} = KL\left ( \frac{1}{|V|}\sum_{j \in |V|}\sum_{i \in P} \mathbf{p_i}(p_i | v_j), \mathbf{U} \right) + \sum_{i \in P} \min_{j \in P,\\ i\neq j} \exp{-d(p_i, p_j)},
\end{equation}

where $\mathbf{U}$ is the uniform distribution and $d$ the euclidean distance.
The first part computes the Kullback-Leibler divergence of the output class distribution and a uniform distribution~\citep{ORCA}.
It prevents the collapse of the output distribution to collapse to only one or few prototypes in early stages of the training by distributing the datapoints over all prototypes.
In cases where information of the prior class distribution is available, one can substitute $\mathbf{U}$ by the known prior $\mathbf{P}$.
The second part increases the distance between the pairwise closest prototypes to separate the prototypes more sharply.
It especially pushes the prototypes of unlabeled classes away from the one of labeled classes such that they get some pseudo-labels assigned during training. 

\section{Additional Experiments}

\subsection{Number of Class Estimation}
\label{app:number_of_class_estimate}
In many real-world settings, the assumption that the number of classes is known beforehand may not be true.
For this reason, we perform an additional experiment where we estimate the number of classes and use the estimated number as the number of prototypes.

To estimate the number of classes of the unlabeled node set $V_u$, we follow the procedure of~\citet{n_class_estimate}.
The training set is divided into a subset of classes for training $Y_t$ and probe classes $Y_p$.
The model is trained on the subgraph $V_t$ of nodes containing only classes from $Y_t$. 
The probing node set $V_p$ is further divided into an anchor set $V_a$ and a validation set $V_v$.
We run a constrained $k$-means~\citep{k-means} on $V_p \cup V_u$, with the constrain that nodes of $V_a$ are forced to match their ground truth label, while considering nodes of $D_v$ as unlabeled.
This procedure is repeated for $k \in \{0, \dots K_{max}\}$ values of $k$ on $V_p \cup V_u$. 
The final number of categories is chosen based on two quality measures. 
First, we measure the labeled clustering quality on the validation probe set $V_v$ by average clustering accuracy~\citep{n_class_estimate}.
Second, we measure the cluster quality of the unlabeled data $V_u$ by silhouette score~\citep{sill-score}. 
Afterwards, we select $\hat{k}= \frac{1}{2}(k_v^* + k_u^*)$, where $k_v^*$ is the $k$ that maximized the average clustering accuracy on $V_v$ and $k_u^*$ is the $k$ that maximized the silhouette score on $V_u$.
To finally determine the number of classes in $V_u$, we run $k$-means one more time with $k=\hat{k}$ and remove all clusters that contain less than $5\%$ of the data. 
The number of remaining clusters is set to the number of classes in $V_u$.

We conduct experiments where we estimate the number of classes and use it as the number of prototypes for POWN on Photo and OGB-arXiv.
The results are presented in Table~\ref{tab:n_class_estimation}.

\begin{table}[!ht]
    \centering
    \begin{tabular}{l|rrr}
    \toprule
     Dataset & Estimate per fold & $|Y|$ & \multicolumn{1}{c}{Error rate in percent}\\
     \midrule
     Photo        &  $\{11, 11, 12, 12\}$   & $8$ & $\{37.5, 37.5, 50.0, 50.0\}$\\
     OGB-Arxiv    &  $\{51, 51, 51, 51, 51\}$  &  $40$ & $\{27.5, 27.5, 27.5, 27.5,27.5\}$\\
     \bottomrule
    \end{tabular}
    \caption{The estimated numbers of new classes, the true number of classes, and the relative error for each fold of the datasets.}
    \label{tab:n_class_estimation}
\end{table}

We repeat our main experiment with the estimated numbers of new classes for each fold instead of the true number to see how sensitive the model performance is towards this parameter.
The results for these experiments are presented in Table~\ref{tab:estimated_class_results}.

\begin{table}[!ht]
    \centering
    \begin{tabular}{l|ccc|ccc}
    \toprule
       &   \multicolumn{3}{c|}{Photo} & \multicolumn{3}{c}{ OGB-arXiv}\\
    \midrule
    & All & Known & New & All & Known & New \\
    \midrule
    POWN + known number of classes& $\mathbf{71.27}$ & \underline{$91.05$} & \dashuline{$70.76$} & $55.51$ & \underline{$71.24$} & \dashuline{$56.11$}\\
    POWN + estimated number of classes & $70.05$ & $88.43$ & $70.76$ & $\mathbf{55.65}$ & $67.80$ & $53.56$ \\
    \bottomrule
    \end{tabular}
    \caption{The results for POWN with an estimated number of classes.}
    \label{tab:estimated_class_results}
\end{table}

The results with the estimated classes show a small drop of less than $4\%$ on all performance measures, which shows that POWN can also be applied in settings where no prior information on the number of classes is available.

\subsection{Analysis of Edge weights in Label Propagation}
\label{app:lp_weights}

\begin{figure}
    \centering
    \includegraphics[width=0.6\textwidth]{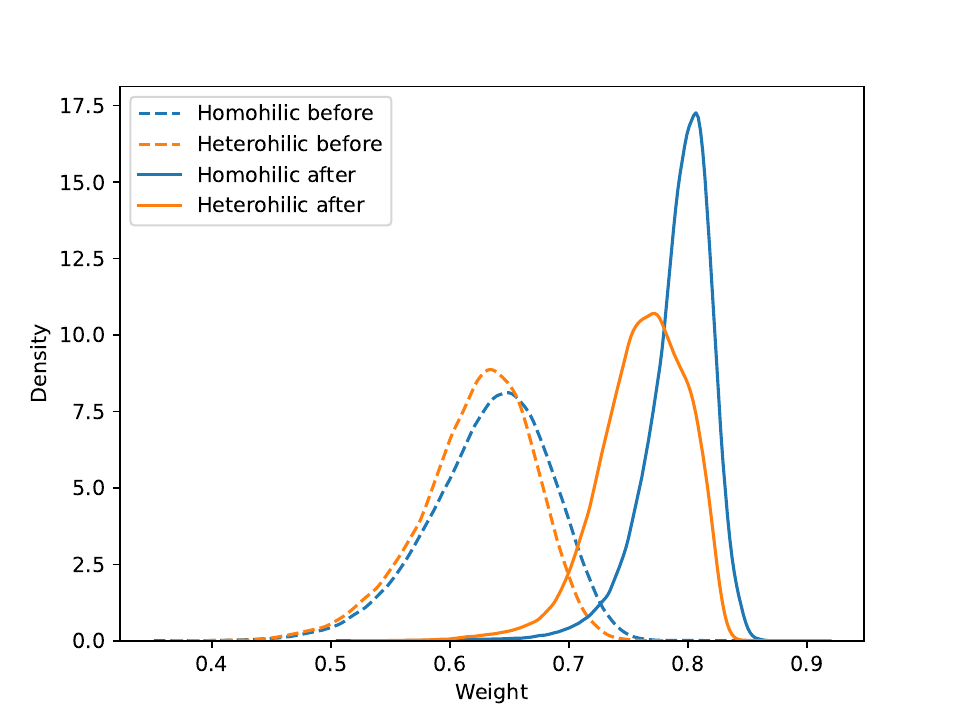}
    \caption{The distribution of edge weights before and after training, separated by homophilic and heterophilic edges.}
    \label{fig:edge_weight}
\end{figure}

To analyze the dynamics of our weights, we compare the distribution of the edge weights of Equation~\ref{eq:weights}.
We trained POWN as described in Section~\ref{sec:procedure} on the first class fold of the Photos dataset.
The edge weights before and after training separated by homophilic and heterophilic edges are shown in Figure~\ref{fig:edge_weight}.

The figure shows that the mean edge weights increase and the standard deviation decreases.
This shows that the embeddings are closer to their prototypes. 
Furthermore, the difference of the means before training is $0.0113$, and after training it is $0.0299$, which is an increase of $164,6\%$.
Therefore, the edge weights are more likely to inhibit the label flow between heterophilic edges and facilitate the label flow of homophilic edges.

\subsection{Prototype as Mean}
\label{app:proto_mean}
We conducted experiments, where we treated the prototype as the weighted mean of its class embedding~\citep{GPN, Geometer,OWAGL}.
Instead of being a trainable parameter, the prototype is the weighted mean of the embeddings with ground truth label or pseudo-label $y$:

\begin{equation*}
    p_c = \sum \limits_{z_i \in V_p, y_i=y} \frac{\exp{(\alpha_y)}}{\sum \limits_{z_k \in V_p, y_k=y} \exp{(\alpha_k)}} z_i \,,
\end{equation*}

where $\alpha_j$ is the page-rank value of node $v_j$ to give a higher weight for nodes in central positions of the graph, and $z_i$ is the embedding of $v_i$.
Furthermore, the prototype loss of POWN is substituted by a labeled contrastive loss that pushes embeddings with the same label together:

\begin{equation*}
    \mathcal{L}_{S/P}= -\frac{1}{|\mathcal{P}(v)|} \sum_{z^+ \in \mathcal{P}(v)} \log \left(\frac{\exp{(z^T z^+ / \tau)}}{\sum_{z^- \in V_l} \exp{(z^T z^- / \tau)}}\right) \,,
\end{equation*}
where $v$ is a node, $z=f(v)$ the embedding of $v$ by a model $f$, $\mathcal{P}(v)$ the set of positive samples for $v$, and $\tau$ some smoothness parameter.
In this case, the positive set is the set of nodes with the same label, which can be either given as part of the gold standard or from a pseudo label.

In pre-experiments, we analyzed the difference between using the trainable prototypes POWN versus weighted means with a contrastive loss.
The results can be found in Table~\ref{tab:mean_results}.
The experiments show that this prototype representation performs worse for our problem. 
We did not follow up on this direction since the mean variant performed worse than simply using a GCN. 
However, we acknowledge that there are further optimizations possible such as potentially a better choice of hyperparameter values.

\begin{table*}[!ht]
    \centering
    \begin{tabular}{l|rrrr}
    \toprule
    Model & \multicolumn{1}{c}{Cora} & \multicolumn{1}{c}{CiteSeer} & \multicolumn{1}{c}{Photo} & \multicolumn{1}{c}{Computers}\\
    \midrule
    POWN-mean & $43.01$ / $10.82$ / $51.99$ & $35.30$ / $15.88$ / $38.90$ & $46.12$ / $10.64$ / $59.69$ & $39.34$ / \:$8.13$ / $53.61$\\
    POWN   & $61.28$ / $87.95$ / $63.29$ & $56.15$ / $75.45$ / $57.29$ & $71.27$ / $91.05$ / $70.76$ & $71.33$ / $82.45$ / $65.61$\\
    \bottomrule
    \end{tabular}
    \caption{Comparison of POWN as an end-to-end model to POWN-mean. POWN-mean computes the prototypes as the mean of the embeddings per class.}
    \label{tab:mean_results}
\end{table*}

\subsection{Sensitivity Analysis of the Temperature parameters}
\label{sec:temp_sens}

\add{
Additionally to the sensitivity analysis in the main paper, we analyze the effect of the temperature parameter of the supervised loss $\tau_S$ and the pseudo-label loss $\tau_P$.}

\begin{figure}[!ht]
     \centering

     \begin{subfigure}[b]{0.8\textwidth}
         \centering
         \includegraphics[width=\textwidth]{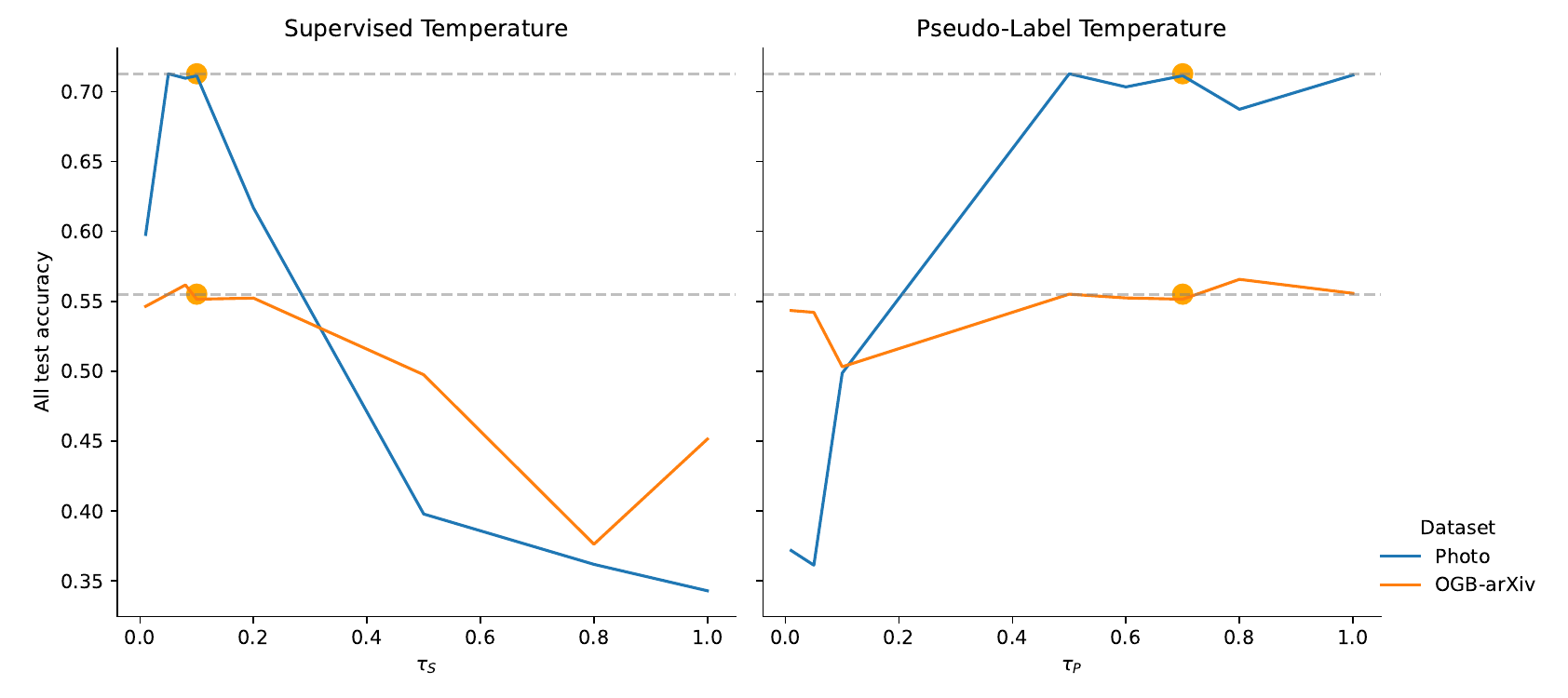}
         \caption{Analysis of the temperature parameters for the supervised loss $\tau_S$ and the pseudo-label loss $\tau_P$ for a range between $0.01$ and $1$. The orange dot marks the temperature used on the experiments.}
         \label{fig:local_temp_analysis}
     \end{subfigure}
     \hfill
     \begin{subfigure}[b]{0.8\textwidth}
         \centering
         \includegraphics[width=\textwidth]{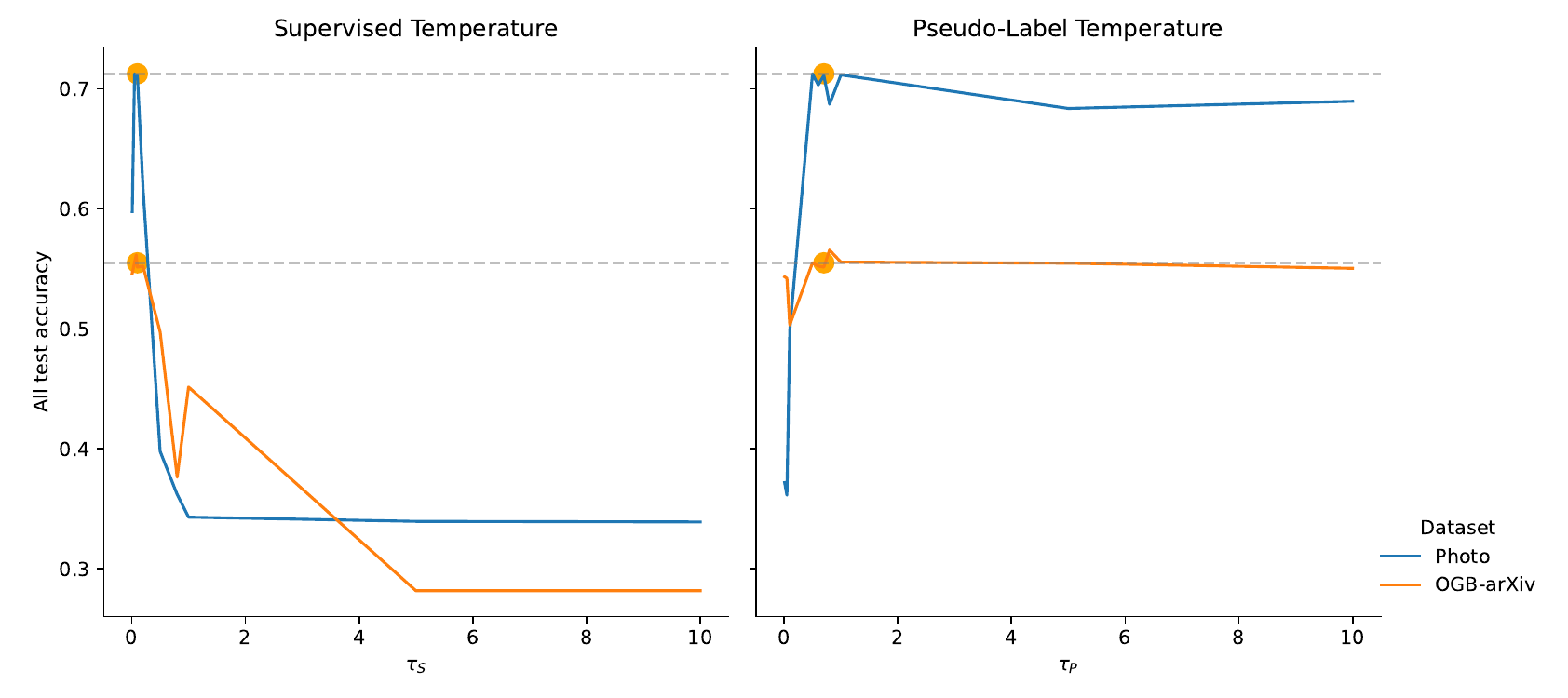}
         \caption{Analysis of the temperature parameters for the supervised loss $\tau_S$ and the pseudo-label loss $\tau_P$ for a range between $0.01$ and $10$. The orange dot marks the temperature used on the experiments.}
         \label{fig:global_temp_analysis}
     \end{subfigure}
     \caption{Ablation study on the temperature parameters of the supervised and pseudo-label loss.}
\end{figure}

We use the datasets Photo and OGB-arXiv. 
We fix all hyperparameters to the results of our hy\-per\-pa\-rameter search and variate $\tau_S \in \{0.01,0.05, 0.08, 0.1, 0.2, 0.5, 0.8, 1.0, 5.0, 10.0\}$ and $\tau_P \in \{0.01, 0.05, 0.1, 0.5 ,0.6, 0.7, 0.8, 1.0, 5.0, 10.0\} $. 
The results are presented in Figures \ref{fig:local_temp_analysis} and \ref{fig:global_temp_analysis}.

\add{The sensitivity analysis of our temperature parameters shows that they were chosen in a reasonable range, which is close to optimum.
The temperature parameter determines how much randomness is used in the contrastive loss.
If the supervised temperature becomes too large, we do not use any label information since all prototype assignments become random.
This is similar to not using the supervised loss at all.
For the pseudo label temperature, we have an increase in the beginning. Since our pseudo-labels have some noise, the smoothing of the temperature parameters accounts for the uncertainty.
For high temperatures, we observe that on Photo the performance converges to some sub-optimal value while staying at the value of our main results on OGB-arXiv.
Therefore, the pseudo-label assignment is more important for Photo than for OGB-arXiv, which is consistent with the results of our ablation study (see Table~\ref{tab:ablation_study}.
However, the plot suggests that there could exist a better value for $\tau_P$ on OGB-arXiv, which is larger than the one we used.
}

\section{Dataset Details}
\subsection{Dataset Node Split}
\label{app:node_split}
We split the nodes for each dataset in the following way.
For Cora and CiteSeer, we use the Planetoid split~\citep{Planetoid}.
For Photo and Computers, we follow \citet{PitfallsGNNEval} and use a fixed, random Planetoid split with $20$ train, $500$ validation, and $1,000$ test nodes per class.
We use the default temporal splits for OGB-arXiv and Reddit2.
For OGB-arXiv, the training nodes are all papers published until $2017$, the validation nodes are from $2018$, and the test nodes are all papers published since $2019$~\citep{ogb-benchmarks}.
Reddit2 has been sampled in October and November $2014$.
For Reddit2, the first $21$ days are used for training.
From the remaining days, $30\%$ are used for validation and the rest as test set~\citep{GraphSage}.

\subsection{Dataset Fold Details}
\label{sec:ds_details}
We split the datasets into $3$ to $5$ folds, such that each fold contains a subset of the classes.
We ensure that each fold contains roughly $20\%$ of the classes and at least $2$ classes.
The resulting distributions of the number of classes into the folds can be seen in Table~\ref{tab:folds}.

\begin{table}[!ht]
    \centering
    \begin{tabular}{l|rrrrr|rr}
    \toprule
    Dataset & 1 & 2 & 3 & 4 & 5 & Train F. & $|C_k|$\\
    \midrule
    Cora & 2 & 2 & 3 & - & - & 1 & 2-3\\
    CiteSeer & 2 & 2 & 2 & - & - & 1 & 2 \\
    Photo & 2 & 2 & 2 & 2 & - & 2 & 4 \\
    Computers & 2 & 2 & 2 & 2 & 2 & 3 & 6\\
    OGB-Arxiv & 8 & 8 & 8 & 8 & 8 & 3 & 24\\
    Reddit2 & 8 & 8 & 8 & 8 & 9 &  3 & 24-25\\
    \bottomrule
    \end{tabular}
    \caption{Number of classes per fold, the number of folds used for training, and the number of known classes $|C_k|$ the model is trained on in each fold.}
    \label{tab:folds}
\end{table}

\section{Reproducibility and Hyperparameter}
\label{appendix:hyperparams}

We trained all models using the Adam optimizer~\citep{adam} with early stopping on the validation accuracy.
All runs of $k$-means are initialized by $k$-means++~\citep{k-means++}.
We set the patience to $30$ for the small datatasets, \ie Cora, CiteSeer, Photo, Computers, and $20$ for the larger datasets, \ie OGB-arXiv and Reddit2. 
Furthermore, we used the neighborhood sampler~\citep{GraphSage} with a batch size of $4,096$ and an upper bound of $128$ for the number of one-hop and $32$ for the number of two-hop neighbors for OGB-arXiv and Reddit2.

For the GCN encoder, which is used by all methods that require an encoder in our experiments, we choose the hyperparameters as presented in Table~\ref{tab:gcn_hp}.

\begin{table*}[!ht]
    \centering
    \begin{tabular}{l|rrrrr}
        \toprule 
         Dataset & Layer & Hidden dim & Dropout & Learning rate & weight decay  \\
         \toprule
         Cora & $2$ & $128$ & $0.4$ & $0.01$ & $0.001$ \\
         CiteSeer & $2$ & $256$ & $0.8$ & $0.01$ & $0.01$\\
         Photo & $2$ & $64$ & $0.8$ & $0.01$ & $0.001$\\
         Computers & $2 $& $64$ & $0.8$ & $0.01$ & $0.001$\\
         OGB-arXiv & $3$  & $1024$ & $0.6$ & $0.01$ & $0.001$ \\
         Reddit2 & $2$ & $128$ & $0.2$ & $0.001$ & $0.0$ \\
         \bottomrule
    \end{tabular}
    \caption{Chosen hyperparameter values for the GCN encoder.}
    \label{tab:gcn_hp}
\end{table*}

For OpenWGL, we separately tuned the hidden dimension in $\{16,32,64\}$.
The best hidden dimension was $32$ for all datasets, except Reddit2, where we used $16$.

POWN has a weight for each loss and a temperature for the prototype losses, \ie the supervised and pseudo-label loss.
We set the temperature for the supervised loss $\tau_S = 0.1$ and for the pseudo-labels as $\tau_P = 0.7$, following \citet{OpenCon}.
The loss weights $\lambda$, $\mu$, $\nu$, and $\kappa$ are tuned by Bayesian hyperparameter optimization.
We performed $1,000$ optimization steps for the small datasets, Cora, CiteSeer, Amazon and Photo, and $100$ steps for the large datasets OGB-ArXiv and Reddit2.
We select the hyperparameters with the highest average validation accuracy over all folds for each dataset. 
Each loss weight is tuned on a range between $0$ and $1$, with a uniform prior distribution.
For the label propagation of POWN, we perform $2$ iterations and only keep the labels with the lowest $10\%$ entropy.
The final parameters are presented in Table~\ref{tab:pown_final_hp}.

\begin{table*}[!ht]
    \centering
    \footnotesize
    \begin{tabular}{l|rrrrr}
    \toprule
      Dataset & \multicolumn{1}{c}{$\lambda$} & \multicolumn{1}{c}{$\mu$} & \multicolumn{1}{c}{$\nu$} & \multicolumn{1}{c}{$\kappa$} & \multicolumn{1}{c}{$q$} \\
         \midrule 
         Cora & $0.596017$ & $0.652459$ & $0.763453$ & $0.208553$  & $0.333999$\\
         CiteSeer & $0.550021$ & $0.238629$ & $0.951837$ & $0.021996$ & $0.525537$\\
         Photo & $0.863386$ & $0.308698$ & $0.461507$ & $0.101025$ & $0.882899$\\
         Computers & $0.988548$ & $0.274850$ & $0.362032$ & $0.174794$ & $0.635788$\\
         OGB-arXiv & $0.549624$  & $0.670594$ & $0.262712$ & $0.139899$ & $0.334155$\\
         Reddit2 & $0.983511$ & $0.829771$ & $0.068015$ & $0.022779$ & $0.915777$ \\
         \bottomrule
    \end{tabular}
    \caption{Final hyperparameter values for the loss weights $\lambda$, $\mu$, and $\nu$, the regularization weight $\kappa$, and the threshold $\gamma$, which determines the subset $V_n$ of our method POWN.}
    \label{tab:pown_final_hp}
\end{table*}

\section{Computing Infrastructure}
The results were computed on a server with two AMD EPYC 9534 64-core Processors, two terabytes of RAM, and one Nvidia H 100 80 GB GPU.

\newpage
\section{Notation Table}

\begin{table}[!ht]
    \centering
    \begin{tabular}{ll}
    \toprule
       Variable  & Definition \\
    \midrule
     $G$    & Graph consisting of nodes and edges \\
     $V$    & All nodes of the graph $G$ \\
     $V_l$  & Subset of labeled nodes with labels from known classes \\
     $V_u$ & Subset of unlabeled nodes from known and new classes\\
     $V_n$ & Subset of unlabeled nodes which are predicted to belong to a new class\\
     $E$    & All edges of the graph $G$ \\
     $Y$ & Set of all classes \\
     $Y_k$ & Set of known classes, where a label is present in the training set \\
     $Y_n$ & Set of new classes, where no label is available during training \\
     $P$ & Set of all prototypes \\
     $P_k$ & Subset of prototypes associated with a known class \\
     $p_i$ & A specific prototype that represents class $i$ \\
     $w_j$ & Edge weight computed by distance between the source node and the closest prototype\\
     $\mathcal{L}_{nll}(V', P')$ & Negative log likelihood loss of the prototype  
     distances, parameterized by a subset  \\ & of the nodes $V'$ and a subset of the prototypes $P'$ \\
     $\mathcal{L}_{S}$ & Supervised loss applied for the labeled classes\\
     $\mathcal{L}_{U}$ & Unsupervised loss applied on all datapoints \\
     $\mathcal{L}_{P}$ & Prototypical loss applied to the prototypes for new classes by using pseudo-labels \\
     $\mathcal{R}$ & Regularization term based on KL divergence and prototype distance\\
     $f$ & Encoder model to obtain an embedding for a node \\
     $x$ & Feature vector associated with a node $v$ \\
     $z$ & Embedding of node $v$ \\
     $y$ & Specific class of a node $v$\\
     $\hat{y}$ & Pseudo label assigned to specific node\\
     $\tilde{z}$ & Corrupted version of the embedding $z$ used for DGI\\
     $d(z_i,z_j)$ & Distance between the embeddings $z_i$ and $z_j$\\
     $\tau$ & Temperature hyperparameter \\
     $D$ & Discriminator for the DGI loss represented by a two layer GCN\\
     $s$ & Summary function required for DGI as mean of the embeddings \\
     $q$ & Ratio of nodes which are considered as a new class \\
     $\gamma$ & The similarity where $q\%$-percent of the labeled nodes are not contained\\
     $\lambda$ & Weight of the supervised loss \\
     $\mu$ & Weight of the unsupervised loss \\
     $\nu$ & Weight of the prototype loss \\
     $\kappa$ & Weight of the regularization term \\

    \bottomrule
    \end{tabular}
    \caption{Variables used throughout the paper along with their meaning.}
    \label{tab:notation}
\end{table}

\end{document}